\documentclass[letterpaper, 10 pt, conference]{ieeeconf}  

\IEEEoverridecommandlockouts                              

\usepackage{amsmath} 

\usepackage{graphicx}
\usepackage{makecell}
\usepackage{threeparttable} 
\usepackage{subcaption}
\usepackage{xcolor}

\usepackage[pagebackref = false,breaklinks,colorlinks,citecolor=green]{hyperref}

\newif\ifproofread
\proofreadfalse

\title{\LARGE \bf
GPS-VIO Fusion with Online Rotational Calibration
}

\author{Junlin Song$^{1}$, Pedro J. Sanchez-Cuevas$^{2}$, Antoine Richard$^{1}$, Raj Thilak Rajan$^{3}$ and Miguel Olivares-Mendez$^{1}$
\thanks{This research was supported by the European Union’s Horizon 2020 project SESAME (grant agreement No 101017258). Dr. Raj Thilak Rajan is partially funded by the European Leadership Joint Undertaking (ECSEL JU), under grant agreement No. 876019, the ADACORSA project, the Sensor AI Lab, under the AI Labs program of Delft University of Technology, and the Moonshot project. 
}
\thanks{$^{1}$ Space Robotics (SpaceR) Research Group, Int. Centre for Security, Reliability and Trust (SnT), University of Luxembourg, Luxembourg.} 
\thanks{$^{2}$Advanced Centre for Aerospace Technologies (CATEC) Seville, Spain.} 
\thanks{$^{3}$Signal Processing Systems, Faculty of Electrical Engineering (EEMCS), 
Delft University of Technology (TUD), Delft.}
}

\begin{document}

\maketitle
\thispagestyle{empty}
\pagestyle{empty}

\begin{abstract}

Accurate global localization is crucial for autonomous navigation and planning. To this end, various GPS-aided Visual-Inertial Odometry (GPS-VIO) fusion algorithms are proposed in the literature. 
This paper presents a novel GPS-VIO system that is able to significantly benefit from the online calibration of the rotational extrinsic parameter between the GPS reference frame and the VIO reference frame. The behind reason is this parameter is observable.
This paper provides novel proof through nonlinear observability analysis.
We also evaluate the proposed algorithm extensively on diverse platforms, including flying UAV and driving vehicle.
The experimental results support the observability analysis and show increased localization accuracy in comparison to state-of-the-art (SOTA) tightly-coupled algorithms. 

\end{abstract}

\begin{keywords}
Sensor Fusion, State Estimation, Kalman Filter
\end{keywords}

\section{Introduction}

The accuracy, robustness and reliability of the pose estimation are essential for the safe autonomous navigation of mobile robots.
In the past few decades, the Global Positioning System (GPS) has been widely used for localization in outdoor scenes, since it offers a robust global localization solution without accumulating drift over time.
However, due to the high-level noise of consumer grade GPS sensors, accurate positioning cannot be typically achieved by using GPS sensors alone.
Moreover, in urban scenes, GPS signals are vulnerable to the interference of the local environment, such as signal occlusions, or bounces due to high-rise buildings, which further degrades the GPS positioning performance.
In these GPS-degraded or -denied scenarios, Visual-Inertial Odometry (VIO) using IMU(Intertial Measurement Unit) and Camera(s), and Simultaneous Localization and Mapping (SLAM) algorithms are conventionally implemented.

VIO algorithms do not suffer from these interruptions, and provide high-precision and high-frequency local state estimation, however, these algorithms also have their inherent drawbacks.
For instance, VIO systems cannot provide long-term drift-free localization and heading.
In \cite{kelly2011visual}, the authors prove that VIO systems have four unobservable Degrees of Freedom (DoF), namely the 3D positions and yaw.
SLAM mitigates this drawback using simultaneous estimation of the localization, the map, along the execution of the loop-closure, and map alignment.
These mechanisms allow SLAM techniques to decrease the localization uncertainty and the long-term drift.
Unfortunately, SLAM techniques demand high computational and memory resources, which limit their applicability.
In general, GPS positioning and visual inertial navigation system can be combined to provide an accurate, high frequency, robust localization and long-term drift-free localization. The fusion of the three sensors involved in this process, GPS, camera and IMU has produced promising results and can achieve locally accurate and globally drift-free localization.

GPS-aided VIO algorithms have been previously proposed in the literature \cite{lee2020intermittent, cioffi2020tightly}. The spatial transformation to couple both the GPS and the VIO reference frame is shown to be unobservable with linear observability analysis \cite{lee2020intermittent, lee2019gps}. However, linearization implies that the derived observability is local and may be unreliable for the original nonlinear system \cite{terrell2001local, tang2008ins}. Thus, it is necessary to revisit the observable property with the tool of nonlinear observability analysis. More specifically, we aim to show in this paper that the rotational extrinsic parameter between GPS reference frame and VIO reference frame is observable.

In this paper, we propose a novel filter-based GPS-VIO system which is specially focused on including a reliable and accurate estimation of the rotation extrinsic between the GPS reference frame and the VIO reference frame. Our key contributions are summarized as:

\begin{itemize}
\item We propose a novel filter-based estimator to fuse GPS measurements and visual-inertial data, and simultaneously estimate the rotational extrinsic parameter between the GPS and VIO frames online.

\item We prove that the rotational extrinsic parameter is observable via nonlinear observability analysis, and support our conclusion with simulations.

\item We evaluate the localization accuracy of the proposed algorithm on multiple public datasets, including small scale flying datasets and large scale driving datasets, and show the superior performance of our algorithm.


\end{itemize}

\section{Related work} \label{Related work}

Sensor fusion of camera and IMU is a well studied topic \cite{huang2019visual, delmerico2018benchmark}. Visual-inertial fusion algorithms can be broadly classified into two categories i.e., optimization-based methods and filter-based methods. Optimization-based methods achieve higher theoretical accuracy, which include VINS-Mono \cite{qin2018vins}, Basalt \cite{usenko2019visual} and ORB-SLAM3 \cite{campos2021orb}. Their high computational cost is a major disadvantage. In contrast, sliding-window  filter-based methods, such as the Multi-State Constraint Kalman Filter (MSCKF) \cite{mourikis2007multi, sun2018robust, geneva2020openvins}, are more resource efficient and achieve comparable accuracy. 

The combination of a camera and an IMU can only generate relative pose estimation, resulting in the unobservability of global position and absolute yaw \cite{kelly2011visual}. 
Therefore, pure VIO systems tend to drift over time \cite{huang2014towards}. 
Recent works have employed GPS measurement to eliminate this drift. These methods can be divided into loosely-coupled methods and tightly-coupled methods. VINS-Fusion is a loosely-coupled approach, which fuses GPS position measurements and output pose of VIO subsystem \cite{qin2019general}. However, the fusion algorithm is unable to improve the VIO subsystem. Therefore, the inner correlations of all measurements are discarded, causing suboptimal localization results. Gomsf is a similar loosely-coupled work \cite{mascaro2018gomsf}.

Tightly-coupled methods fully exploit the complementary merits of multi-sensor data, and are promising to further improve the accuracy. A tightly-coupled estimator based on sliding window optimization is proposed in \cite{yu2019gps}. The rotation between the GPS reference frame and the VIO reference frame is included in the state vector, but the non-synchronization between GPS timestamp and VIO system timestamp is neglected. \cite{cioffi2020tightly} describes another tightly-coupled optimization-based approach. The comparative experiments with VINS-Fusion have demonstrated that tightly-coupled methods are superior to loosely-coupled methods. However, the transformation between GPS reference frame and VIO reference frame is not estimated in \cite{cioffi2020tightly}.  The closest to our work is \cite{lee2020intermittent}, which is a tightly-coupled estimator based on MSCKF. 
The extrinsic parameters between the GPS reference frame and the VIO reference frame are inserted into the state during initialization, however, marginalized after all states are transformed from the VIO reference frame to the GPS reference frame \cite{lee2020intermittent}. 

Consequently, the approach of \cite{lee2020intermittent} does not estimate the extrinsic parameters between GPS-VIO online, as they show the extrinsic parameters are unobservable with linear observability analysis. However, linear observability analysis maybe unreliable for a nonlinear system. A locally observable system is sure to be globally observable, but a locally unobservable system maybe globally observable \cite{tang2008ins, wu2012observability, hermann1977nonlinear}. Our main contribution in this work is to point out that rotational extrinsic parameter is globally observable using nonlinear observability analysis. This novel observability conclusion is similar to our recent accepted work \cite{song2023gps}, termed as GPS-VWO.
The difference between GPS-VIO and GPS-VWO lies in the different kinematic equations, with the former driven by IMU while the latter driven by wheel odometer. As the reference frame of VIO is gravity aligned, the rotational extrinsic parameter of GPS-VIO only has yaw component. Unlike GPS-VIO, the rotational extrinsic parameter of GPS-VWO is 3DoF in general. To analyze the observability of extrinsic parameter, Lie derivative is employed for GPS-VIO, like GPS-VWO \cite{song2023gps}, considering the nonlinearity of this system.

The unavoidable errors caused by imposing fixed extrinsic parameters after GPS-VIO initialization lead to miss-calculations of the fusion algorithms in long distances. Without online calibration, the estimation error of rotational extrinsic parameter at the start will deteriorate the localization accuracy, especially when the GPS noise is relatively large. \cite{yu2019gps, cao2022gvins, boche2022visual} adopt explicitly online calibration of the rotational extrinsic parameter to improve localization accuracy. To simplify the state estimation complexity, \cite{boche2022visual} disables the online estimation once the rotational extrinsic parameter is converged. However, neither of them provide a theoretical observability analysis. In this paper, we prove the rotational extrinsic parameter is observable; hence, including it in the state vector is a promising and theoretically guaranteed mean to improve the accuracy of the state estimator.

\section{Problem Formulation}

\subsection{Reference frames and Notation} 
\begin{figure}[htbp]
  \centering
  \includegraphics[width=0.25\textwidth]{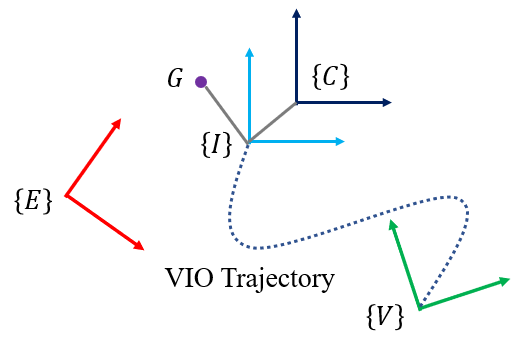}
  \caption{Coordinate systems, similar as Fig. 1a in \cite{song2023gps}.}
  \label{Frames}
\end{figure}

The coordinate systems used in this work are shown in the Fig. \ref{Frames}. $\{ E\} $ represents the East-North-Up (ENU) coordinate system, which is the reference frame of GPS position measurements. An arbitrary GPS measurement can be chosen as the origin of $\{ E\} $ frame. $\{ V\} $ is the reference frame of the VIO system. After the initialization of the VIO system, the orientation of this coordinate system is gravity aligned. $\{ I\} $ and $\{ C\} $ represent IMU coordinate frame and camera coordinate system, respectively. $G$ is the position of the antenna of the GPS receiver.

We use the notation ${}^V\left(  \bullet  \right)$  to represent a quantity in the coordinate frame $\{ V\} $. The position of point $I$ in the frame $\{ V\} $ is expressed as ${}^V{p_I}$. The velocity of frame $\{ I\} $ in the frame $\{ V\} $ is expressed as ${}^V{v_I}$. Furthermore, we use quaternion to represent the attitude of rigid body \cite{trawny2005indirect}. ${}_V^Iq$ represents the orientation of frame $\{ I\} $ with respect to frame $\{ V\} $, and its corresponding rotation matrix is given by ${}_V^IR$. Similar notations also apply to the other reference frames. ${\left[  \bullet  \right]_ \times }$ denotes the skew symmetric matrix corresponding to a three-dimensional vector, and ${\left[  \bullet  \right]^T}$ is used to represent the transpose of a matrix. 

\subsection{Classical MSCKF-based VIO structure}

According to \cite{lee2020intermittent, geneva2020openvins}, the classic MSCKF-based VIO algorithm usually defines the following states
\begin{equation} \label{MSCKF}
    \begin{array}{l}
    x = {\left[ {\begin{array}{*{20}{c}}
    {x_I^T}&{x_{{c_1}}^T}& \cdots &{x_{{c_N}}^T}
    \end{array}} \right]^T}\\
    {x_I} = {\left[ {\begin{array}{*{20}{c}}
    {{}_V^I{q^T}}&{{}^Vp_I^T}&{{}^Vv_I^T}&{{}^Vp_f^T}&{b_\omega ^T}&{b_a^T}
    \end{array}} \right]^T}\\
    {x_{{c_i}}} = {\left[ {\begin{array}{*{20}{c}}
    {{}_V^{{I_i}}{q^T}}&{{}^Vp_{{I_i}}^T}
    \end{array}} \right]^T}
    \end{array}
\end{equation}
where ${}^V{p_f}$ represents feature position, $f$,  in the VIO reference frame $\{ V\} $. To make the presentation concise, only one feature point is described here. ${b_\omega }$ and ${b_a}$ are the biases of the IMU angular velocity and the linear acceleration measurements, respectively. ${x_I}$ indicates the current IMU state. ${x_{{c_i}}}$  is obtained by extracting the first two quantities of $x_I$ at different image times. Then the state of the whole MSCKF system $x$ can be constructed by augmenting $N$ historical ${x_{{c_i}}}$ in ${x_I}$.

After successful initialization and setting appropriate initial value and covariance for $x$, the VIO system follows the Kalman filter pipeline. IMU measurements are used for the propagation of ${x_I}$. Whenever a new image is received, $x$ is augmented with the pose clone of the current ${x_I}$ and the visual constraints between multiple pose clones are utilized to update the state. For more details of this part, we refer interested readers to Open-VINS \cite{geneva2020openvins}.

\subsection{GPS Measurement Update} \label{GPS_Update}

Assuming that the first GPS position measurement as the origin of the $\{ E\} $ frame, the subsequent GPS measurements are denoted as ${}^E{p_G}$. Each GPS observation can be formulated as\footnote{Measurement equation (\ref{GPS}) is used here just for the convenience of the observability analysis. In the implementation, we adopt the interpolation measurement equation as \cite{lee2020intermittent}.}
\begin{equation} \label{GPS}
    \begin{split}
    z &= {}^E{p_G} + n_{gps} = {}^E{p_V} + {}_V^ER{}^V{p_G} + n_{gps}\\
     &= {}^E{p_V} + {}_V^ER\left( {{}^V{p_I} + {}_V^I{R^T}{}^I{p_G}} \right) + n_{gps}
    \end{split}
\end{equation}
where $n_{gps}$ is a  white Gaussian noise. ${}^I{p_G}$ is the position of point $G$ in the IMU frame $\{ I\} $. This paper assumes that this quantity is known, since ${}^I{p_G}$ can be obtained from CAD model or calibrated before the system runs. ${}_V^IR$ and ${}^V{p_I}$ are quantities expressed in the VIO reference frame $\{ V\} $. ${}^E{p_V}$ and ${}_V^ER$ are the transformations between frame $\{ V\} $ and frame $\{ E\} $. Since these two frames are gravity aligned, we can simply use the yaw angle to parameterize the rotation matrix between them. Therefore, ${}_V^ER$ can be expressed as
\begin{equation}
    {}_V^ER = \left[ {\begin{array}{*{20}{c}}
    {\cos \psi }&{ - \sin \psi }&0\\
    {\sin \psi }&{\cos \psi }&0\\
    0&0&1
    \end{array}} \right]
\end{equation}
where $\psi $ is the relative yaw angle between GPS reference frame and VIO reference frame.

To make the measurement equation usable, $\psi $ and ${}^E{p_V}$ must be known. In \cite{lee2020intermittent}, these are calculated in the initialization stage of the GPS-VIO system and are marginalized later. 
The main difference between our work and theirs is that we will provide a more suitable nonlinear observability analysis in Section \ref{Nonlinear Observability Analysis}, to decide the inclusion of observable quantities to the system state vector for potential online refinement.

To analyze the observability of all the extrinsic parameters between the frame $\{ V\} $ and the frame $\{ E\} $, $\psi$ and ${{}^Ep_V}$ are included in the state vector. Moreover, the time offset between the GPS and IMU should also be modeled for the sake of real world experiments. Thus, the new system state vector then becomes
\begin{equation}
    x = {\left[ {\begin{array}{*{20}{c}}
    {x_I^T}&{x_{{c_1}}^T}& \cdots &{x_{{c_N}}^T}&\psi &{{}^Ep_V^T} &{{}^It_G}
    \end{array}} \right]^T}
    \label{eq:nsv}
\end{equation}
where $\psi$ and ${{}^Ep_V}$ represent the interested extrinsic parameters, and ${{}^It_G}$, is the time offset between the GPS and IMU clock\footnote{As time offset ${{}^It_G}$ calibration is not the focus of this work, it is ignored in following analysis, but considered in real-world experiments to compensate non-synchronization (Section \ref{KAIST EXP}).}. The subset of system state related to the GPS measurement equation is noted as ${x_s}$
\begin{equation}
    {x_s} = {\left[ {\begin{array}{*{20}{c}}
    {{}_V^I{q^T}}&{{}^Vp_I^T}&\psi &{{}^Ep_V^T}
    \end{array}} \right]^T}
\end{equation} The measurement Jacobian $H$ is expressed as
\begin{equation}
    H = \frac{{\partial \tilde z}}{{\partial {{\tilde x}_s}}}
     = \left[ {\begin{array}{*{20}{c}}
    { - {}_V^E R{}_V^I{R^T}{{\left[ {{}^I{p_G}} \right]}_ \times }}&{{}_V^E R}&{{H_\psi {}^V{p_G}}}&{{I_3}}
    \end{array}} \right]
\end{equation}
\begin{equation}
    {H_\psi } = \left[ {\begin{array}{*{20}{c}}
    { - \sin \psi }&{ - \cos \psi }&0\\
    {\cos \psi }&{ - \sin \psi }&0\\
    0&0&0
    \end{array}} \right]
\end{equation}

To keep this paper focused and concise, we omit the description of the GPS-VIO system's initialization, as well as the proper handling of time-offsets among the different sensors, like ${{}^It_G}$. Interested readers can refer to \cite{lee2020intermittent}.

\section{Observability Analysis}

\subsection{Comments on Linear Observability Analysis}

The linear observability analysis of the GPS-VIO system has been investigated previously in \cite{lee2020intermittent} and detailed in \cite{lee2019gps}. However, the unobservable property obtained about extrinsic parameters in these works may be misleading, as they apply linear observability analysis for a typically nonlinear GPS-VIO system. As discussed in Section \ref{Related work}, a locally unobservable system maybe globally observable \cite{tang2008ins, wu2012observability, hermann1977nonlinear}. Moreover, no experiments were performed in \cite{lee2020intermittent, lee2019gps} to validate the observability conclusions regarding the extrinsic parameters. In this work, we employ a more appropriate nonlinear observability analysis for the nonlinear GPS-VIO system to obtain observable property and provide experiments to solidify the analysis.

\subsection{Nonlinear Observability Analysis} \label{Nonlinear Observability Analysis} We now conduct a nonlinear observability analysis, following the standard Lie derivatives method \cite{hermann1977nonlinear}. Removing the IMU bias and pose clones from the state vector (\ref{eq:nsv}), to simplify the formulation, the state becomes
\begin{equation}
   x = \left[ {\begin{array}{*{20}{c}}
   {{}_V^I{q^T}}&{{}^Vv_I^T}&{{}^Vp_I^T}&{{}^Vp_f^T}&\psi &{{}^Ep_V^T}
   \end{array}} \right]
\end{equation} Following immediately, we write the kinematic equations as \begin{equation}
    \left[ {\begin{array}{*{20}{c}}
    {{}_V^I\dot q}\\
    {{}^V{{\dot v}_I}}\\
    {{}^V{{\dot p}_I}}\\
    {{}^V{{\dot p}_f}}\\
    {\dot \psi }\\
    {{}^E{{\dot p}_V}}
    \end{array}} \right] = \underbrace {\left[ {\begin{array}{*{20}{c}}
    {{0_{4 \times 1}}}\\
    g\\
    {{}^V{v_I}}\\
    {{0_{3 \times 1}}}\\
    0\\
    {{0_{3 \times 1}}}
    \end{array}} \right]}_{{f_0}} + \underbrace {\left[ {\begin{array}{*{20}{c}}
    {\frac{1}{2}\Xi \left( {{}_V^Iq} \right)}\\
    {{0_3}}\\
    {{0_3}}\\
    {{0_3}}\\
    0_{1 \times 3}\\
    {{0_3}}
    \end{array}} \right]}_{{f_1}}\omega  + \underbrace {\left[ {\begin{array}{*{20}{c}}
    {{0_{4 \times 3}}}\\
    {{}_V^I{R^T}}\\
    {{0_3}}\\
    {{0_3}}\\
    0_{1 \times 3}\\
    {{0_3}}
    \end{array}} \right]}_{{f_2}}a
\end{equation}where $g$ is denoted as the local gravity vector. ${\omega }$ and ${a}$ are de-biased IMU angular velocity and linear acceleration measurements, respectively. Here, we use the time derivative property of quaternions
\begin{equation}
    \dot q = \frac{1}{2}\Omega \left( \omega  \right)q = \frac{1}{2}\Xi \left( q \right)\omega
\end{equation}

The definition of $\Xi \left( q \right)$  can be found in \cite{trawny2005indirect}.

Next, we list the usable measurement equations. The camera measurement equation is
\begin{equation}
    {h_1}\left( x \right) = {}^C{p_f} = {}_I^CR{}_V^IRp + {}^C{p_I}
    \label{h_1}
\end{equation}
where $p = {}^V{p_f} - {}^V{p_I}$. The norm constraint of the unit quaternion is also considered as a measurement equation
\begin{equation}
    {h_2}\left( x \right) = {}_V^I{q^T}{}_V^Iq - 1 = 0
    \label{h_2}
\end{equation} The measurement equation of the GPS is
\begin{equation}
    {h_3}\left( x \right) = {}^E{p_G} = {}^E{p_V} + {}_V^ER{}^V{p_I}
    \label{h_3}
\end{equation} where without loss of generality, we assume ${}^I{p_G} = {0_{3 \times 1}}$ to simplify the expression.


\subsubsection{Zeroth-Order Lie Derivatives}The zeroth-order Lie derivative of a function is itself.
\begin{equation}
    \begin{array}{l}
    {\pounds^0}{h_1} = {}^C{p_I} + {}_I^CR{}_V^IRp\\
    {\pounds^0}{h_2} = {}_V^I{q^T}{}_V^Iq - 1\\
    {\pounds^0}{h_3} = {}^E{p_V} + {}_V^ER{}^V{p_I}
    \end{array}
\end{equation}The gradients of zeroth-order Lie derivatives with respect to $x$  are
\begin{equation}
    \begin{array}{l}
    \nabla {\pounds^0}{h_1} = \left[ {\begin{array}{*{20}{c}}
    {X_1 }&{{0_3}}&{ - {}_I^CR{}_V^IR}&{{}_I^CR{}_V^IR}&{{0_{3 \times 1}}}&{{0_3}}
    \end{array}} \right]\\
    \nabla {\pounds^0}{h_2} = \left[ {\begin{array}{*{20}{c}}
    {2{}_V^I{q^T}}&{{0_{1 \times 3}}}&{{0_{1 \times 3}}}&{{0_{1 \times 3}}}&0&{{0_{1 \times 3}}}
    \end{array}} \right]\\
    \nabla {\pounds^0}{h_3} = \left[ {\begin{array}{*{20}{c}}
    {{0_{3 \times 4}}}&{{0_3}}&{{}_V^ER}&{{0_3}}&{{H_\psi }{}^V{p_I}}&{{I_3}}
    \end{array}} \right]
    \end{array}
\end{equation} where $X$ represents a quantity that does not need to be computed explicitly, as it does not affect the observability analysis.

\subsubsection{First-Order Lie Derivatives}
The first-order Lie derivative of $h_1$ with respect to  $f_0$ is computed as
\begin{equation}
    \pounds_{{f_0}}^1{h_1} = \nabla {\pounds^0}{h_1} \bullet {f_0} = - {}_I^CR{}_V^IR{}^V{v_I}
\end{equation} The gradient of $\pounds_{{f_0}}^1{h_1}$  with respect to $x$  is \begin{equation}
    \nabla \pounds_{{f_0}}^1{h_1} = \left[ {\begin{array}{*{20}{c}}
    {{X_2}}&{ - {}_I^CR{}_V^IR}&{{0_3}}&{{0_3}}&{{0_{3 \times 1}}}&{{0_3}}
    \end{array}} \right]
\end{equation} The first-order Lie derivative of $h_1$  with respect to  $f_1$ is computed as
\begin{equation}
    \pounds_{{f_1}}^1{h_1}  = \nabla {\pounds^0}{h_1} \bullet {f_1} = \frac{1}{2}{X_1}\Xi \left( {{}_V^Iq} \right)
\end{equation} where the gradient of $\pounds_{{f_1}}^1{h_1}$  with respect to $x$  is
\begin{equation}
    \nabla \pounds_{{f_1}}^1{h_1} = \left[ {\begin{array}{*{20}{c}}
    {X_3 }&{{0_{9 \times 3}}}&{X_4 }&{-X_4 }&{{0_{9 \times 1}}}&{{0_{9 \times 3}}}
    \end{array}} \right]
\end{equation}

The first-order Lie derivative of $h_3$  with respect to $f_0$  is computed as
\begin{equation}
    \pounds_{{f_0}}^1{h_3} = \nabla {\pounds^0}{h_3} \bullet {f_0} = {}_V^ER{}^V{v_I}
\end{equation} and the gradient of $\pounds_{{f_0}}^1{h_3}$  with respect to $x$  is
\begin{equation}
    \nabla \pounds_{{f_0}}^1{h_3} = \left[ {\begin{array}{*{20}{c}}
    {{0_{3 \times 4}}}&{{}_V^ER}&{{0_3}}&{{0_3}}&{{H_\psi }{}^V{v_I}}&{{0_3}}
    \end{array}} \right]
\end{equation}

\subsubsection{Observability analysis}\hfill\\
By stacking the gradients of previously calculated Lie derivatives together, the following observability matrix is constructed
\begin{equation}
    \begin{array}{c}
    {\cal O} = \left[ {\begin{array}{*{20}{c}}
    {\nabla {\pounds^0}{h_1}}\\
    {\nabla {\pounds^0}{h_2}}\\
    {\nabla {\pounds^0}{h_3}}\\
    {\nabla \pounds_{{f_0}}^1{h_1}}\\
    {\nabla \pounds_{{f_1}}^1{h_1}}\\
    {\nabla \pounds_{{f_0}}^1{h_3}}
    \end{array}} \right]
     = \\
    \left[ {\begin{array}{*{20}{c}}
    {{X_1}}&{{0_3}}&{ - {}_V^CR}&{{}_V^CR}&{{0_{3 \times 1}}}&{{0_3}}\\
    {2{}_V^I{q^T}}&{{0_{1 \times 3}}}&{{0_{1 \times 3}}}&{{0_{1 \times 3}}}&0&{{0_{1 \times 3}}}\\
    {{0_{3 \times 4}}}&{{0_3}}&{{}_V^ER}&{{0_3}}&{{H_\psi }{}^V{p_I}}&{{I_3}}\\
    {{X_2}}&{ - {}_V^CR}&{{0_3}}&{{0_3}}&{{0_{3 \times 1}}}&{{0_3}}\\
    {{X_3}}&{{0_{9 \times 3}}}&{{X_4}}&{ - {X_4}}&{{0_{9 \times 1}}}&{{0_{9 \times 3}}}\\
    {{0_{3 \times 4}}}&{{}_V^ER}&{{0_3}}&{{0_3}}&{{H_\psi }{}^V{v_I}}&{{0_3}}
    \end{array}} \right]
    \end{array}
\end{equation}

Adding the fourth column to the third column, ${\cal O}$  becomes
\begin{equation}
    \begin{array}{l}
    {\cal O} = 
    \left[ {\begin{array}{*{20}{c}}
    {{X_1}}&{{0_3}}&{{0_3}}&{{}_V^CR}&{{0_{3 \times 1}}}&{{0_3}}\\
    {2{}_V^I{q^T}}&{{0_{1 \times 3}}}&{{0_{1 \times 3}}}&{{0_{1 \times 3}}}&0&{{0_{1 \times 3}}}\\
    {{0_{3 \times 4}}}&{{0_3}}&{{}_V^ER}&{{0_3}}&{{H_\psi }{}^V{p_I}}&{{I_3}}\\
    {{X_2}}&{ - {}_V^CR}&{{0_3}}&{{0_3}}&{{0_{3 \times 1}}}&{{0_3}}\\
    {{X_3}}&{{0_{9 \times 3}}}&{{0_{9 \times 3}}}&{ - {X_4}}&{{0_{9 \times 1}}}&{{0_{9 \times 3}}}\\
    {{0_{3 \times 4}}}&{{}_V^ER}&{{0_3}}&{{0_3}}&{{H_\psi }{}^V{v_I}}&{{0_3}}
    \end{array}} \right]
    \end{array}
\end{equation}

${}_V^ER$ in the third column can be used to eliminate ${{H_\psi }{}^V{p_I}}$  in the fifth column and ${I_3 }$  in the sixth column. Thus, ${\cal O}$ can be reduced to
\begin{equation}
    \begin{array}{l}
    {\cal O} = 
    \left[ {\begin{array}{*{20}{c}}
    {{X_1}}&{{0_3}}&{{0_3}}&{{}_V^CR}&{{0_{3 \times 1}}}&{{0_3}}\\
    {2{}_V^I{q^T}}&{{0_{1 \times 3}}}&{{0_{1 \times 3}}}&{{0_{1 \times 3}}}&0&{{0_{1 \times 3}}}\\
    {{0_{3 \times 4}}}&{{0_3}}&{{}_V^ER}&{{0_3}}&{{0_{3 \times 1}}}&{{0_3}}\\
    {{X_2}}&{ - {}_V^CR}&{{0_3}}&{{0_3}}&{{0_{3 \times 1}}}&{{0_3}}\\
    {{X_3}}&{{0_{9 \times 3}}}&{{0_{9 \times 3}}}&{ - {X_4}}&{{0_{9 \times 1}}}&{{0_{9 \times 3}}}\\
    {{0_{3 \times 4}}}&{{}_V^ER}&{{0_3}}&{{0_3}}&{{H_\psi }{}^V{v_I}}&{{0_3}}
    \end{array}} \right]
    \end{array}
\end{equation}

The sixth column corresponds to the translation part of the extrinsic parameter between frame $\{ V\} $ and frame $\{ E\}$.
This column is not full rank, so the translation part is unobservable.
Finally, we analyze the rotation part of the extrinsic parameter.
Let us focus on  ${{H_\psi }{}^V{v_I}}$ in the fifth column.
The fifth column cannot be eliminated by other columns and is full rank in general case.
So the rotational extrinsic parameter is observable.
It is worth noting that the rank of fifth column can become zero if zero velocity motion incurs, more specifically, $x$ and $y$ direction. Therefore, horizontal velocity excitation affects the observability of the rotational extrinsic parameter.

\section{Results}   

We develop the proposed algorithm based on Open-VINS \cite{geneva2020openvins}, which is a state-of-the-art VIO framework. When GPS information is usable, the system state including the rotational extrinsic parameter between the GPS reference frame and the VIO reference frame is updated via Section \ref{GPS_Update}.
Since \cite{lee2020intermittent} is not open-sourced, we have implemented our own version by following their paper.
In the results presented here, our implementation \cite{lee2020intermittent} is referred as \textbf{“GPS-VIO-fixed”}. The prefix ``fixed'' comes from the fact that the spatial transformation is marginalized after the initialization of the system.
While our algorithm continues to calibrate the rotational extrinsic parameter online after initialization.

First, we design a simulation environment to verify the observability conclusion.
Then, the proposed algorithm is evaluated on two public datasets.
One is the small-scale EuRoC dataset \cite{burri2016euroc}, which has seen extensive use in the VIO research community.
Noisy GPS measurements are simulated by adding Gaussian noise to the groundtruth position. It is featured by UAV flying.
Another one is the large-scale KAIST dataset with real GPS measurements in challenging urban scenes \cite{jeong2019complex}. It is featured by vehicular driving.
The path length and GPS noise of each selected KAIST sequence is longer than 7km and larger than 6m, respectively.

\subsection{Validation of the Observability Analysis} 

To verify our observability proof, we build a simulation environment based on Open-VINS \cite{geneva2020openvins}.
The groundtruth trajectory of MH\_01\_easy in EuRoC dataset is used to generate simulated multi-sensor data, including 400Hz IMU, 10Hz image and 10Hz GPS.
The noise of the GPS sensor is approximated by applying multivariate Gaussian noises with a standard deviation of 0.2m on the positions.

\begin{figure}[htbp]
  \centering
    \begin{subfigure}[t]{0.23\textwidth}
        \centering
        \includegraphics[width=\textwidth, height=\textwidth]{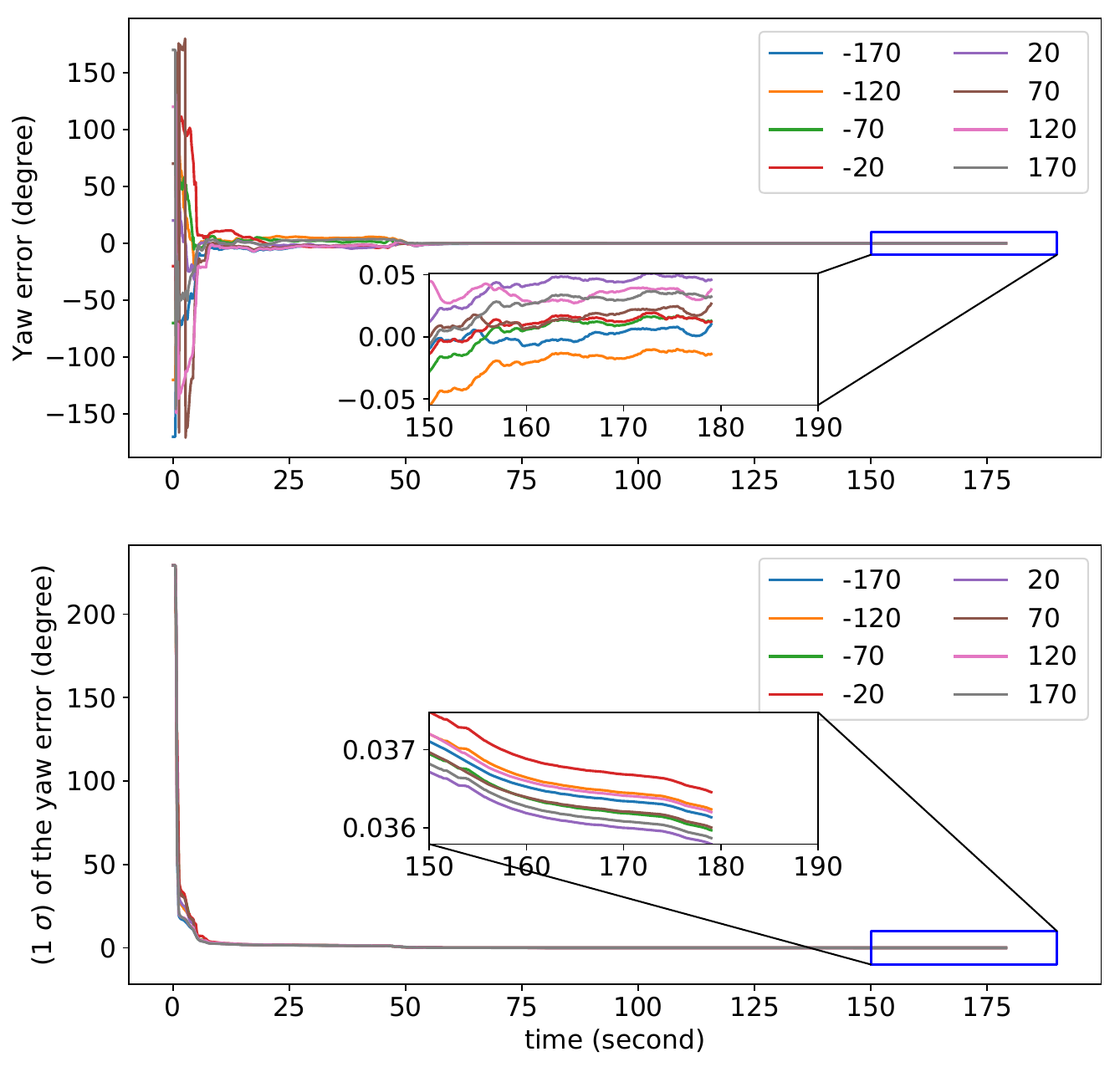}
        \caption{MH\_01\_easy}
        \label{Yaw_merge_euroc}
    \end{subfigure}
    \hfill
    \begin{subfigure}[t]{0.23\textwidth}
        \centering
        \includegraphics[width=\textwidth, height=\textwidth]{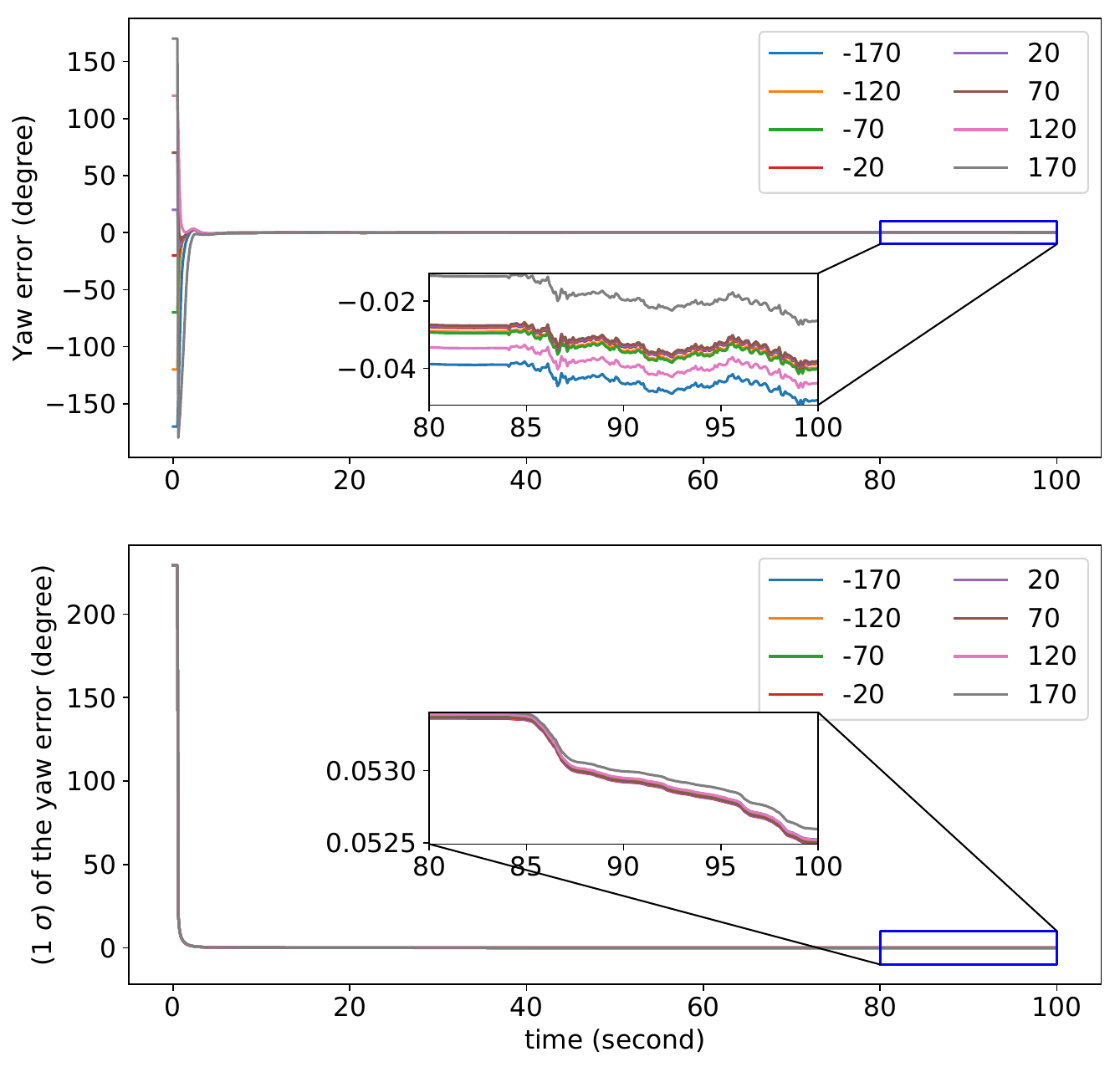}
        \caption{Urban39}
        \label{Yaw_merge_kaist}
    \end{subfigure}
  \caption{Top: $\psi $ convergence over time respect to different initial guesses. Bottom: One standard deviation (1 $\sigma $) of $\psi $.}
  \label{Yaw_merge_euroc_kaist}
\end{figure}

To verify the convergence capability of discovered observable quantity, the calibration of the rotation extrinsic parameter is performed with different initial guesses.
We start with an error of 20 degrees and add 50 degrees increment until we reach 170 degrees, we then reiterate with negative angles from -20 to -170 degrees.
Fig. \ref{Yaw_merge_euroc} shows the convergence of the yaw error.
Between 21s to 45s, the convergence of yaw error reaches to steady state because of the stationary motion status.
As mentioned in Section \ref{Nonlinear Observability Analysis}, zero velocity motion leads to the unobservability of rotation extrinsic parameter.
At other times, there exist velocity excitation. Before 21s, the motion space near the starting point is relatively small compared to GPS noise. After 45s, the moving distance exceeds 10m, which is far greater than GPS noise.

Fig. \ref{Yaw_merge_euroc} shows one standard deviation (1 $\sigma $) of the yaw error. Initial one standard deviation of the yaw error is set to 4 rad, considering the largest initial yaw error is close to $\pi$ rad.
The estimation of the yaw error consistently converges to near zero with small uncertainty, and the convergence process is robust to the relatively large initial error.

Apart from UAV trajectory, we also repeat the above steps with the planer vehicular trajectory of Urban39 in KAIST dataset. Larger GPS noise and practical GPS noise characteristic are considered. The vertical noise is set to twice the horizontal noise. The GPS noise is defined as
\begin{equation}
    n_{gps} \sim {\cal N}\left( {{0_{3 \times 1}},{diag(1, 1, 4)}} \right)
\end{equation}

Fig. \ref{Yaw_merge_kaist} shows the convergence results with vehicular trajectory. Both yaw error and its corresponding uncertainty consistently converges to near zero, even for the near $\pi$ rad initial error. All these results from Fig. \ref{Yaw_merge_euroc_kaist} support that the rotational extrinsic parameter is observable.

\subsection{EuRoC dataset} 

There are 11 sequences in EuRoC dataset. Each sequence is classified into easy, medium or hard according to the level of difficulty for the VIO algorithms.
Image and IMU data are available at 20Hz and 200Hz respectively, and the groundtruth position and orientation are provided at 200Hz.
We test all sequences to verify the convergence of the rotational extrinsic parameter between the GPS reference frame and the VIO reference frame.
Similarly to the previous experiment, the simulated GPS measurements are obtained by adding Gaussian noise ($\sigma  = 0.2m$) to the groundtruth position.
The GPS frequency is sampled to be 20Hz.

For the initialization of our GPS-VIO system, it is assumed that we do not have any accurate initial estimation for $\psi $.
And the initial value of $\psi $ is naively set as $\hat \psi  = 0$. ${}^E{\hat p_V}$ is set as the first GPS measurement received after successful VIO initialization.
The groundtruth of $\psi $ is acquired by querying the groundtruth orientation value at the initialization time.

\begin{figure}[htbp]
  \centering
    \begin{subfigure}[t]{0.23\textwidth}
        \centering
        \includegraphics[width=\textwidth, height=\textwidth]{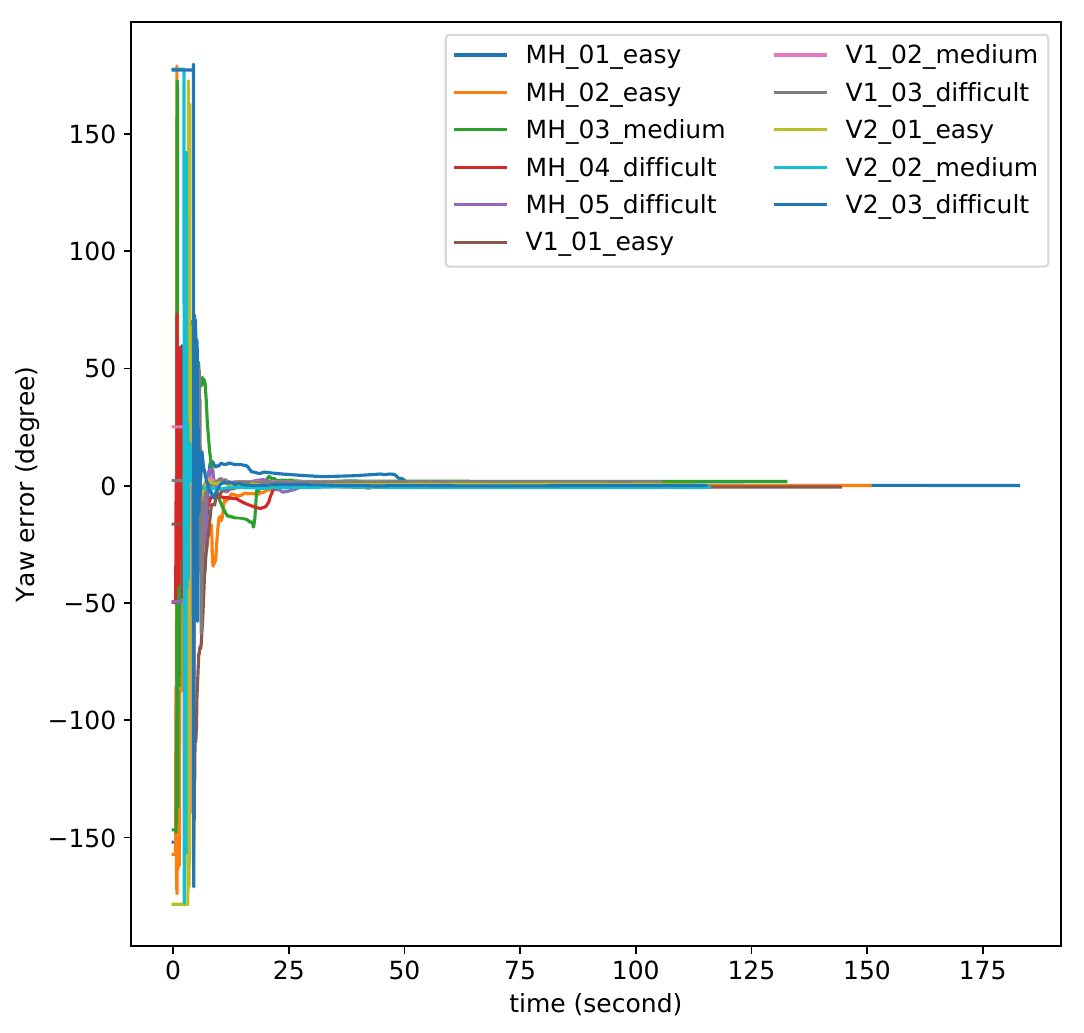}
        \caption{}
        \label{EuRoC Yaw error}
    \end{subfigure}
    \hfill
    \begin{subfigure}[t]{0.23\textwidth}
        \centering
        \includegraphics[width=\textwidth, height=\textwidth]{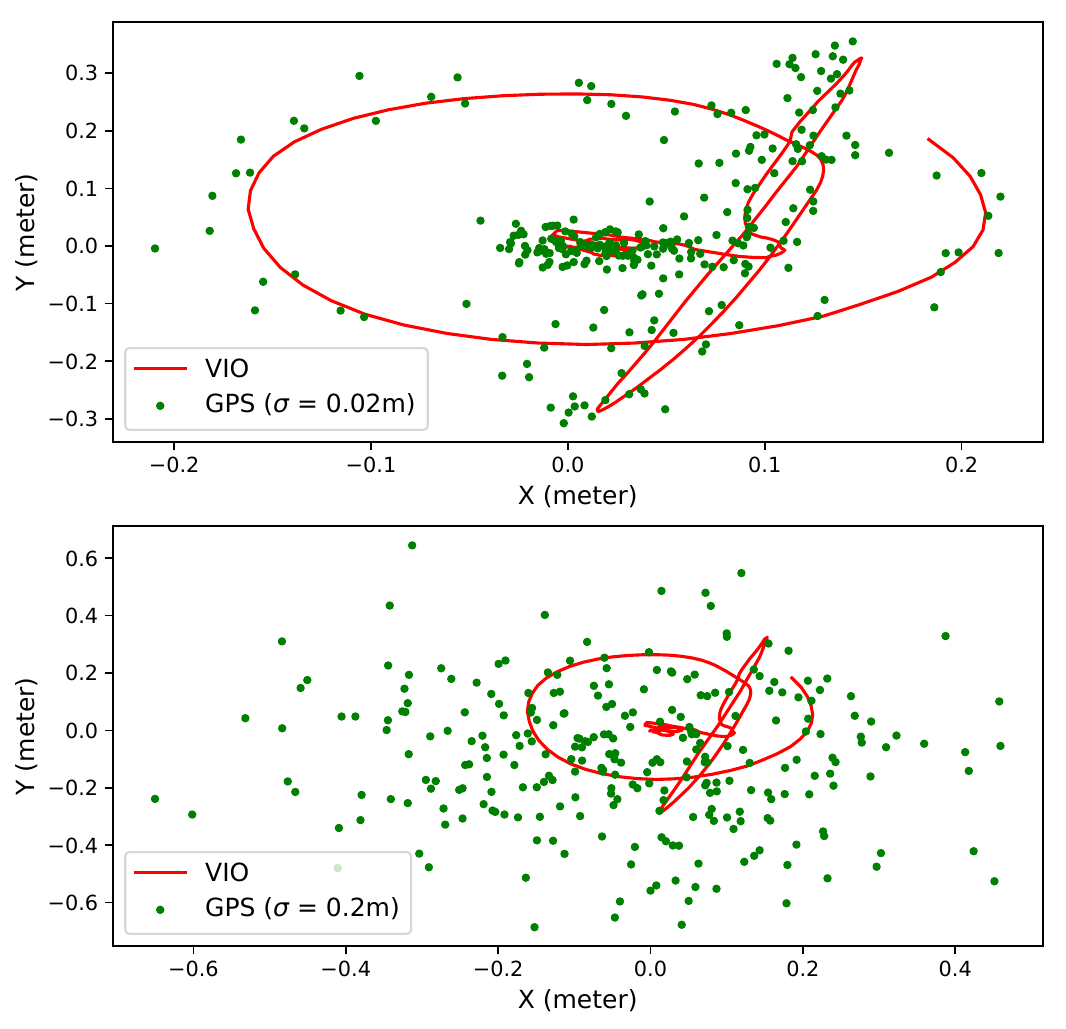}
        \caption{}
        \label{traj_align}
    \end{subfigure}
  \caption{(a) $\psi $ convergence over time. (b) Horizontal view of aligned trajectory with different level of GPS noise.}
\end{figure}

Fig. \ref{EuRoC Yaw error} shows the convergence of $\psi $ over time. The range of the initial yaw error is $\left[ { - {{178.47}^ \circ },{{177.67}^ \circ }} \right]$.
Estimation error $\left( {\hat \psi  - \psi } \right)$ of each sequence approaches to near zero quickly and perfectly.
Results verify the observability of $\psi $.


Table~\ref{table_euroc} shows the Absolute Trajectory Error (ATE) of the different algorithms on all the sequences.
We include the results for GPS positioning, optimization-based VIO  (SVO2.0 \cite{forster2016svo}), filter-based VIO (Open-VINS \cite{geneva2020openvins}), two variants of the tightly-coupled optimization-based GPS-VIO approach \cite{cioffi2020tightly}, GPS-VIO-fixed \cite{lee2020intermittent} and our proposed algorithm.
As \cite{cioffi2020tightly} relies on manually setting initial rotational extrinsic parameter, we provide two variants: initializing $\psi $ as zero as ours, or initializing $\psi $ as groundtruth.
Our approach outperforms other state-of-the-art competitors on most sequences because of online rotational calibration.
Regarding the first three sequences, we achieve less but close accuracy compared to the second variant of \cite{cioffi2020tightly}. The possible reason is that the VIO subsystem of \cite{cioffi2020tightly}, SVO2.0 \cite{forster2016svo}, performs better than our VIO subsystem, Open-VINS \cite{geneva2020openvins}, in the first three sequences. 
However, SVO2.0 suffers from relatively naive VIO initialization strategy \cite{rpg_svo_pro_open_issue} for most sequences.

\begin{table}[ht]
\caption{ATE (meter) Comparison with the SOTA
on the EuRoC Dataset. The ATE of GPS trajectory is 0.347m.}
\label{table_euroc}
\begin{center}
\begin{tabular}{|c|c|c|c|c|c|c|}
\hline
\textbf{\makecell{ID}} & \textbf{\makecell{VIO \cite{forster2016svo}}} & \textbf{\makecell{VIO \cite{geneva2020openvins}}} & \textbf{\makecell{A}} &
\textbf{\makecell{B}} & \textbf{\makecell{C}} & \textbf{\makecell{Ours}} \\
\hline
MH01 &  0.064 & 0.084 & 0.137 &	\textbf{0.031} &	0.114 &	0.036 \\ \hline
MH02 &	0.052 &	0.086 &	0.110 &	\textbf{0.036} &	0.126 &	0.040 \\ \hline
MH03 &	0.118 &	0.124 &	0.119 &	\textbf{0.048} &	0.174 &	0.062 \\ \hline
MH04 &	0.203 &	0.169 &	0.292 &	0.068 &	0.080 &	\textbf{0.061} \\
\hline
MH05 &	0.240 &	0.200 &	0.312 &	0.056 &	0.176 &	\textbf{0.049} \\
\hline
V101 &	0.064 &	0.054 &	0.0 &	0.041 &	0.039 &	\textbf{0.037} \\
\hline
V102 &	0.082 &	0.046 &	0.312 &	0.048 &	0.050 &	\textbf{0.037} \\
\hline
V103 &	0.066 &	0.048 &	0.365 &	0.068 &	0.091 &	\textbf{0.041} \\
\hline
V201 &	0.085 &	0.041 &	0.106 &	0.038 &	0.098 &	\textbf{0.035} \\
\hline
V202 &	0.111 &	0.040 &	0.123 &	0.046 &	0.042 &	\textbf{0.033} \\
\hline
V203 &	0.156 &	0.067 &	0.154 &	0.098 &	0.073 &	\textbf{0.044} \\
\hline
\end{tabular}
\end{center}
\begin{tablenotes}
   \item $^{1}$ A: results of \cite{cioffi2020tightly} by initializing $\psi $ as zero.
   \item $^{2}$ B: results of \cite{cioffi2020tightly} by initializing $\psi $ as groundtruth.
   \item $^{3}$ C: results of GPS-VIO-fixed \cite{lee2020intermittent} by initializing $\psi $ through Section IV in \cite{lee2020intermittent}, which suffers from relatively large GPS noise (see Fig. \ref{traj_align}). The initialization distance is set as 2m.
   \item $^{4}$ Ours: results of proposed method by initializing $\psi $ as zero.
\end{tablenotes}
\end{table}

\subsection{KAIST dataset} \label{KAIST EXP} 

The KAIST datasets are collected in highly complex urban environments.
It is very challenging to achieve high-precision localization in these environments using consumer grade sensors.
Because many moving objects exist in the streets and dense high-rise buildings corrupt GPS signals. 
Image, IMU and GPS of KAIST datasets are received at 10Hz, 100Hz and 5Hz respectively.

\begin{figure}[htbp]
  \centering
    \begin{subfigure}[t]{0.23\textwidth}
        \centering
        \includegraphics[width=\textwidth, height=\textwidth]{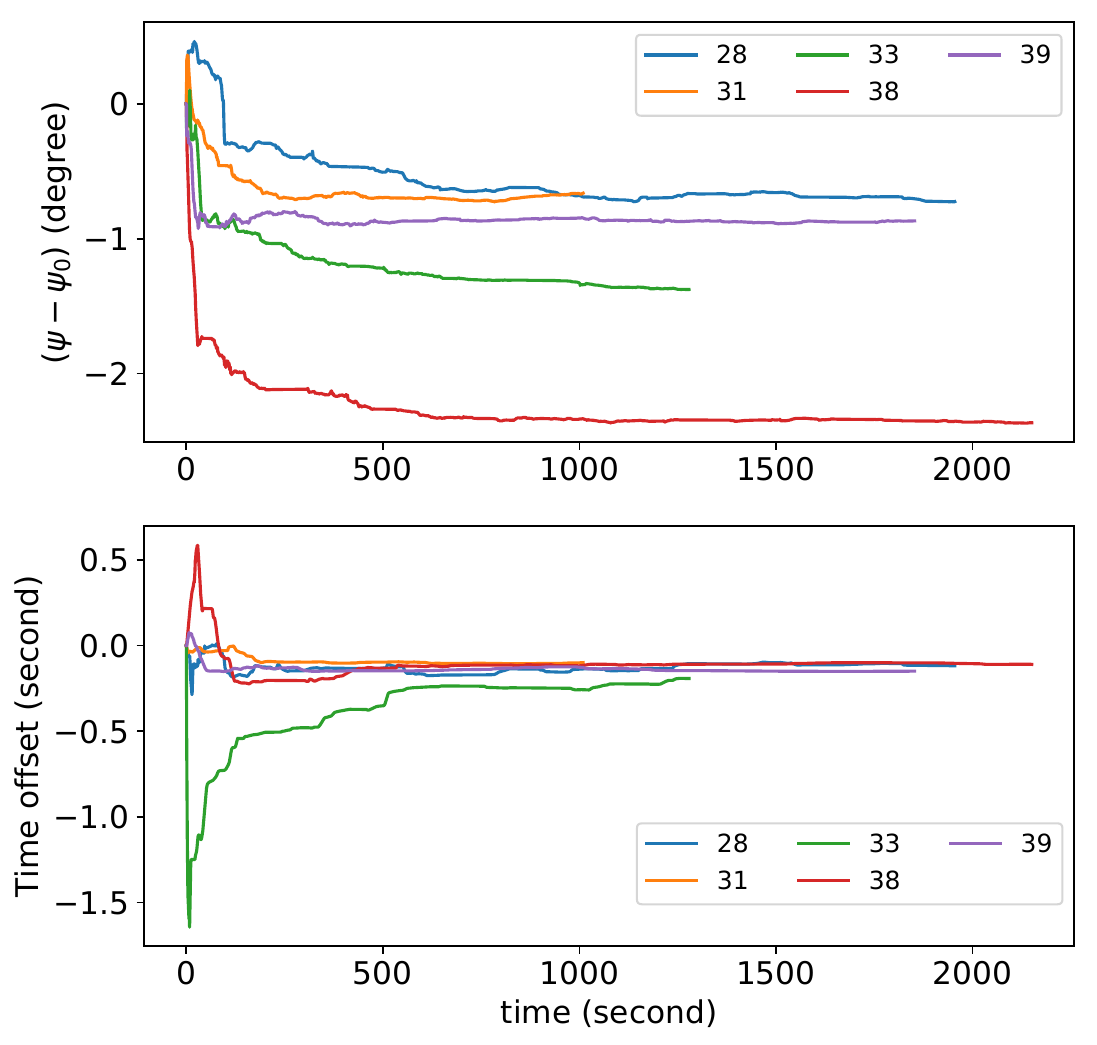}
        \caption{}
        \label{Yaw convergence and time offset}
    \end{subfigure}
    \hfill
    \begin{subfigure}[t]{0.23\textwidth}
        \centering
        \includegraphics[width=\textwidth, height=\textwidth]{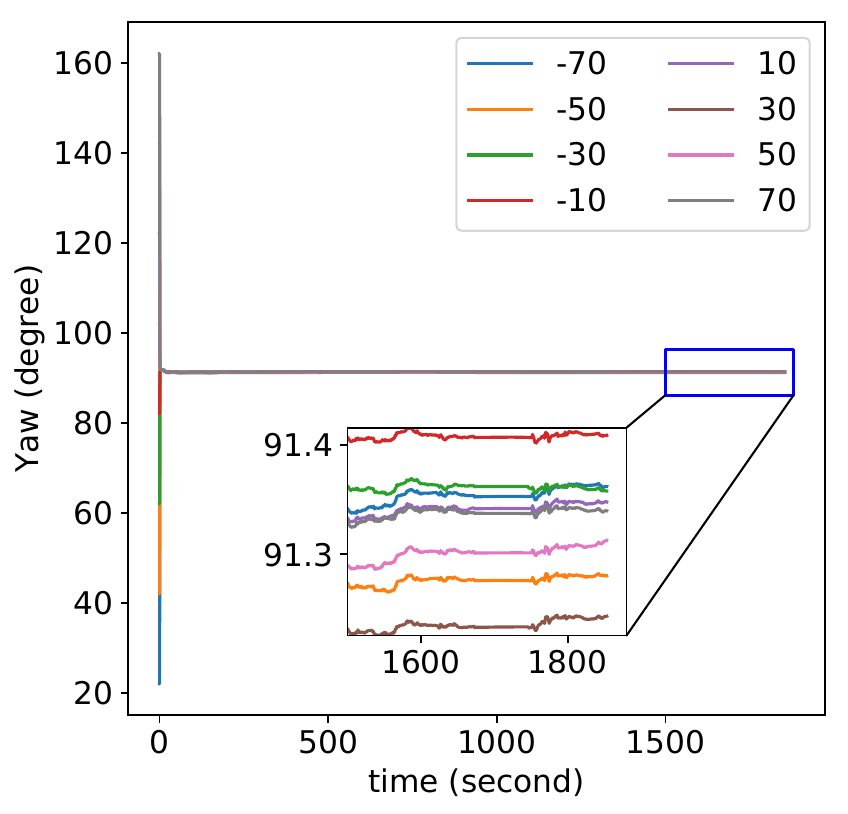}
        \caption{}
        \label{Yaw_repeat_kaist}
    \end{subfigure}
  \caption{(a) Top: $\left( {\psi  - {\psi _0}} \right)$ convergence over time. Bottom: Calibration results of the time offset between GPS and IMU. (b) $\psi $ convergence over time respect to different initial values. The labels of legend represent different perturbation values.}
\end{figure}

We refer to the initialization algorithm of \cite{lee2020intermittent} to obtain the initial $\psi $ and ${}^E{p_V}$. The initialization distance is set as 20m.
${}^E{p_V}$ is fixed after initialization and only $\psi $ is estimated online.
As the groundtruth orientation in the GPS reference frame is unavailable (see Section 4.3 in \cite{jeong2019complex}),
$\left( {\psi  - {\psi _0}} \right)$ is plotted in Fig. \ref{Yaw convergence and time offset} to show the convergence trend over time.
${\psi _0}$ is the initial value of $\psi $.
The deviation from the initial value is less than ${2.5^ \circ }$ for each sequence.

Although the calibration of time offset between GPS and IMU, ${{}^It_G}$, is not the focus of this paper, we still need to deal with it carefully. It has a negative impact on the localization accuracy without proper handling, especially when different sensor clocks are not hardware-synchronized \cite{lee2020intermittent}. Fig. \ref{Yaw convergence and time offset} also shows the time offset calibration results, which are initialized from 0s. The average final converged values is $-0.13\pm0.03$s.

Fig. \ref{Yaw_repeat_kaist} shows the repeatability of yaw calibration for different initial values, with Urban39 dataset. These initial values are obtained by adding different perturbation to $\psi _0$. The range of perturbation is $\left[ { - {{70.0}^ \circ },{{70.0}^ \circ }} \right]$.

We evaluate the ATE of GPS positioning, VIO (Open-VINS \cite{geneva2020openvins}), GPS-VIO-fixed and our proposed algorithm. Results are summarized in TABLE \ref{table_kaist}. Our algorithm provides the highest localization accuracy. VIO suffers from drift issue from long trajectory. Moreover, the scale information of VIO system is unobservable when the vehicle undergoes constant acceleration motion \cite{martinelli2011vision, wu2017vins}. These issues can be solved by fusing GPS measurements once GPS-VIO system is successfully initialized (see Urban33 in TABLE \ref{table_kaist}).

\begin{table}[ht]
\caption{ATE (meter / degree) Comparison with the SOTA
on the KAIST Dataset. $-$ means trajectory divergence.}
\label{table_kaist}
\begin{center}
\begin{tabular}{|c|c|c|c|c|c|c|}
\hline
\textbf{\makecell{ID}} & \textbf{\makecell{Path\\ len(km)}} & \textbf{\makecell{GPS}} & \textbf{\makecell{VIO \cite{geneva2020openvins}}} &
\textbf{\makecell{GPS-VIO-\\ fixed \cite{lee2020intermittent}}} & \textbf{\makecell{Ours}} \\
\hline
28 & 11.5 &	8.66 & 10.78 / 1.44 & 7.71 / 1.75 & \textbf{4.67 / 1.42} \\ \hline
31 & 11.4 &	7.26 & 76.87 / 1.58 & 6.85 / 1.62 & \textbf{5.56 / 1.55} \\ \hline
33 & 7.6 &	8.95 & $-$ & 7.77 / 2.90 & \textbf{4.94 / 1.27} \\ \hline
38 & 11.4 &	7.09 & 7.53 / 1.26 & 5.53 / 1.25 & \textbf{3.86 / 1.22} \\ \hline
39 & 11.1 &	6.43 & 8.73 / 1.93 & 5.50 / 1.48 & \textbf{2.63 / 1.24} \\ \hline
\end{tabular}
\end{center}
\end{table}

\section{Conclusion}

This paper presents a novel tightly-coupled filter-based GPS-VIO algorithm which can benefit from the online estimation of the rotational extrinsic parameter between the GPS and the VIO reference frame.
The proposed algorithm is able to refine the rotational calibration, thus improve the localization performance.
The novel study on the observability of extrinsic parameter demonstrates that nonlinear observability analysis is more comprehensive and profound than linear observability analysis, for a nonlinear system. It is advised to validate the unobservable property derived from linear observability analysis in simulations. In future, we will investigate if we can obtain better localization results by formulating the estimation algorithm directly with the GNSS raw observations \cite{cao2022gvins, lee2022tightly}.









\bibliographystyle{ieeetr}
\bibliography{bib}

\end{document}